\documentclass[11pt]{article}

\usepackage[preprint]{acl}

\usepackage{times}
\usepackage{latexsym}
\usepackage{amsmath}
\usepackage{booktabs}
\usepackage{multirow}
\usepackage{array}
 
\usepackage[T1]{fontenc}
\usepackage[utf8]{inputenc}

\usepackage{microtype}

\usepackage{inconsolata}

\usepackage{graphicx}
\usepackage{pifont}
\newcommand{\cmark}{\ding{51}}
\newcommand{\xmark}{\ding{55}}
\title{One-Eval: An Agentic System for Automated and Traceable LLM Evaluation}

\author{
 \textbf{Chengyu Shen\textsuperscript{1 *}},
 \textbf{Yanheng Hou\textsuperscript{2 *}},
 \textbf{Minghui Pan\textsuperscript{3 *}},
 \textbf{Runming He\textsuperscript{1}},
 \textbf{Zhen Hao Wong\textsuperscript{1}},
 \\
 \textbf{Meiyi Qiang\textsuperscript{1}},
 \textbf{Zhou Liu\textsuperscript{1}},
 \textbf{Hao Liang \textsuperscript{1,4}},
 \textbf{Peichao Lai\textsuperscript{1}},
 \\
 \textbf{Zeang Sheng\textsuperscript{1}},
 \textbf{Wentao Zhang\textsuperscript{1,4 \textdagger}},
\\
 \textsuperscript{1}Peking University,
 \textsuperscript{2}Beijing Institute of Technology,
 \\
 \textsuperscript{3}Beijing University of Posts and Telecommunications,
 \textsuperscript{4}Zhongguancun Academy
\\
 \texttt{scuuy05@gmail.com}
}

\begin{document}
\maketitle
\begin{abstract}
Reliable evaluation is essential for developing and deploying large language models, yet in practice it often requires substantial manual effort: practitioners must identify appropriate benchmarks, reproduce heterogeneous evaluation codebases, configure dataset schema mappings, and interpret aggregated metrics. To address these challenges, we present One-Eval, an agentic evaluation system that converts natural-language evaluation requests into executable, traceable, and customizable evaluation workflows. One-Eval integrates (i) NL2Bench for intent structuring and personalized benchmark planning, (ii) BenchResolve for benchmark resolution, automatic dataset acquisition, and schema normalization to ensure executability, and (iii) Metrics \& Reporting for task-aware metric selection and decision-oriented reporting beyond scalar scores. The system further incorporates human-in-the-loop checkpoints for review, editing, and rollback, while preserving sample evidence trails for debugging and auditability. Experiments show that One-Eval can execute end-to-end evaluations from diverse natural-language requests with minimal user effort, supporting more efficient and reproducible evaluation in industrial settings. Our framework is publicly available at \url{https://github.com/OpenDCAI/One-Eval}.
\end{abstract}

\begingroup
\renewcommand\thefootnote{}
\footnotetext{\textsuperscript{*} Equal contribution.}
\footnotetext{\textsuperscript{\textdagger} Corresponding author.}
\endgroup

\section{Introduction}\label{sec:intro}

With the rapid adoption of large language models and multimodal models in industrial systems~\cite{kimiteam2026kimik25visualagentic, bai2025qwen3vltechnicalreport,liang2025dataflow}, model evaluation has become a critical component throughout the model lifecycle, including development, selection, iteration, and pre-deployment validation~\cite{srivastava2023imitationgamequantifyingextrapolating, chang2023surveyevaluationlargelanguage}. Evaluation results are no longer used solely for reporting benchmark scores, but increasingly serve as decision-making signals for model comparison, deployment readiness, and risk assessment~\cite{yang2025qwen3technicalreport,deepseekai2025deepseekv32pushingfrontieropen}. As evaluation objectives grow more diverse and task-specific, existing evaluation workflows struggle to provide sufficient flexibility and usability in practice.

In current mainstream practices, model evaluation typically follows one of two approaches. Users either identify and reproduce task-specific benchmark repositories~\cite{hendrycks2021measuringmassivemultitasklanguage}, manually setting up environments and running scripts, or rely on static evaluation frameworks that require explicit configuration of models, datasets, parameters, and metrics~\cite{eval-harness, 2023opencompass}. While these approaches standardize execution to some extent, they still place a heavy burden on users to discover appropriate benchmarks, construct valid configurations, and interpret results. Such workflows are highly experience-dependent, costly to iterate, and difficult to adapt to evolving evaluation needs.

Meanwhile, agent-based systems have gained significant traction in industrial applications~\cite{yang2024sweagent, kimiteam2026kimik2openagentic}. Prior work has shown that agentic systems can reduce engineering overhead by allowing users to express high-level goals rather than low-level procedures~\cite{yao2023reactsynergizingreasoningacting, luo2025largelanguagemodelagent, mao2025engineeringmultiagentllmsprotocoldriven}. This motivates a rethinking of model evaluation as an agent-driven task, where the core challenge lies not only in executing evaluations, but in transforming abstract evaluation intents into reliable and actionable evaluation pipelines.

However, treating model evaluation as an end-to-end agent-driven process remains underexplored. Existing tools primarily focus on execution and score aggregation, while treating benchmarks and metrics as static configurations. They rarely address higher-level stages such as evaluation intent interpretation, personalized benchmark selection, configuration validation, or result analysis tailored to downstream decisions. As a result, evaluation outputs are often limited to isolated scalar metrics, which are insufficient for supporting real-world industrial decision making.

In this paper, we propose \textbf{One-Eval}, an agentic evaluation framework that transforms natural language evaluation requests into executable, verifiable, and customizable evaluation workflows. One-Eval follows an end-to-end design with three main stages. First, \textit{NL2Bench} interprets natural language requests, decomposes evaluation intents, and retrieves or recommends benchmarks that align with user goals, with support for interactive refinement. Second, automated benchmark resolution and settings completion handle dataset acquisition, dependency management, and configuration validation, reducing manual effort and configuration errors. Third, One-Eval performs metric recommendation and task-oriented report generation, producing structured, decision-support evaluation reports rather than single scalar scores. To ensure reliability, One-Eval incorporates a human-in-the-loop mechanism at key decision points, enabling users to review and refine agent decisions while preserving automation efficiency.

\section{Related Work}

\textbf{Model Evaluation.}
Model evaluation has long been a central topic in natural language processing and has gained renewed importance with the rise of large language models. A wide range of benchmarks have been proposed to assess model capabilities across domains, including mathematical reasoning benchmarks such as GSM8K~\cite{cobbe2021trainingverifierssolvemath} and MATH~\cite{hendrycks2021measuringmathematicalproblemsolving}, and broad knowledge and reasoning benchmarks such as MMLU~\cite{hendrycks2021measuringmassivemultitasklanguage}. In addition, evaluation toolkits such as lm-eval-harness~\cite{eval-harness} and OpenCompass~\cite{2023opencompass} provide standardized interfaces for running benchmarks and aggregating scores. While these frameworks improve evaluation reproducibility, they largely assume predefined tasks, benchmarks, and metrics, leaving users to manually map evaluation goals to concrete evaluation setups.

\textbf{Automation and Agent-Based Systems.}
Agent-based and multi-agent systems have shown strong effectiveness in automating complex, multi-step tasks such as code generation and tool-oriented workflows~\cite{yang2024sweagentagentcomputerinterfacesenable,wu2023autogenenablingnextgenllm}. By decomposing high-level goals into sequential decisions, these approaches reduce manual effort and support iterative refinement. From a structural perspective, model evaluation is also a multi-stage process involving intent interpretation, benchmark selection, execution, and result analysis. However, existing work has largely applied automation to isolated components, rather than treating it as an end-to-end, agent-driven decision process, resulting in fragmented automation support in practice.

\textbf{Personalized Evaluation and Reporting.}
Most existing evaluation studies present results as single or aggregated metrics~\cite{rein2023gpqagraduatelevelgoogleproofqa,zhong2023agievalhumancentricbenchmarkevaluating}, which support standardized comparison but offer limited guidance for practical deployment decisions. Prior work has explored multi-dimensional evaluation to better characterize model behavior~\cite{liang2023holisticevaluationlanguagemodels,srivastava2023imitationgamequantifyingextrapolating}, yet these approaches typically rely on fixed evaluation dimensions and static reporting formats. As a result, evaluation outputs remain weakly aligned with user-specific goals and task requirements. Motivated by these limitations, our work focuses on evaluation requirement modeling, evaluation workflow automation, and task-oriented report generation, enabling an end-to-end evaluation paradigm driven by user objectives.

\section{System Design}

\subsection{Framework Overview}

\begin{figure*}
\centering 
\includegraphics[width=
1.0\textwidth]{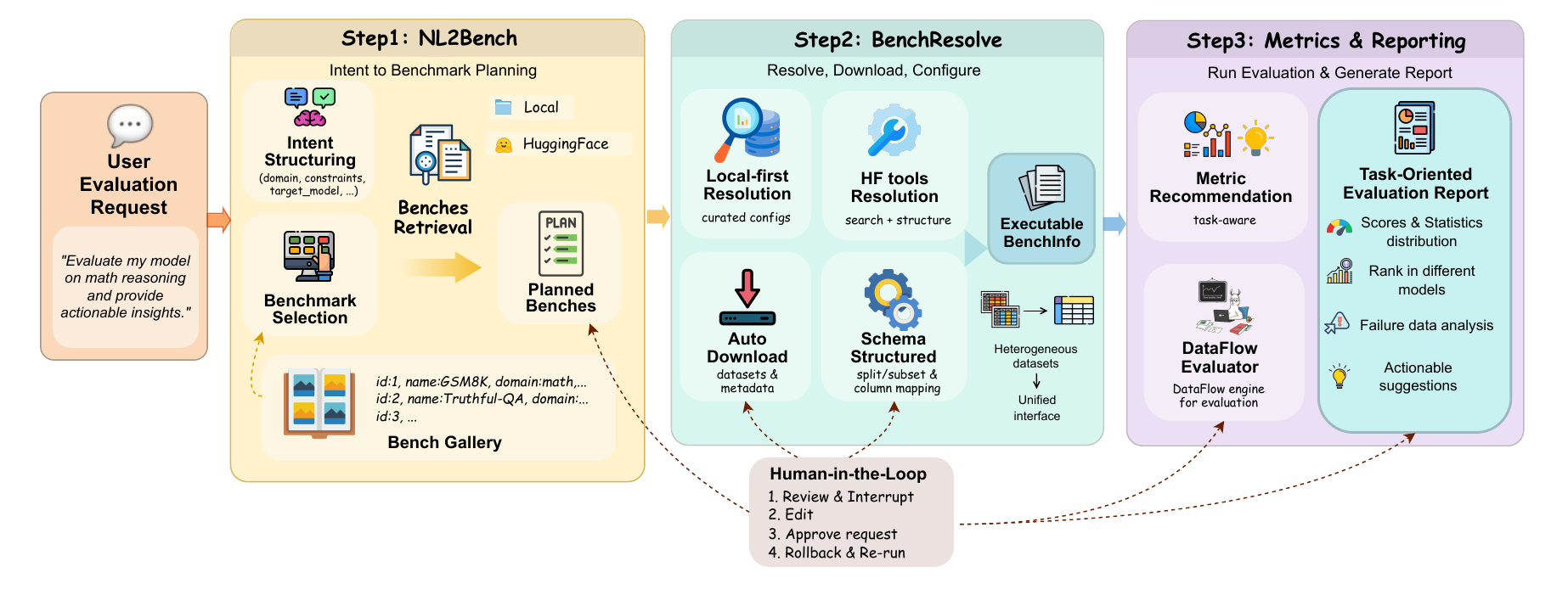} 
\caption{\textbf{One-Eval overview.} One-Eval converts a natural-language evaluation request into an executable \emph{EvalPlan} (NL2Bench), resolves and configures benchmarks by automatic dataset download and schema normalization (BenchResolve), and produces task-aware metrics and a decision-oriented evaluation report (Metrics \& Reporting), with human-in-the-loop refinement at key steps.}
\label{fig:overview}
% \vspace{-4mm}
\end{figure*}

One-Eval is an agentic evaluation framework designed to transform high-level, natural language evaluation requests into executable and verifiable model evaluation workflows. Instead of requiring users to manually identify benchmarks, configure evaluation settings, and interpret results, One-Eval treats model evaluation as an end-to-end decision process driven by user intent.

As illustrated in Figure~\ref{fig:overview}, One-Eval follows a modular, three-stage pipeline. Given a user’s evaluation request expressed in natural language, the framework first interprets the evaluation intent and constructs an appropriate evaluation plan. It then resolves benchmarks and evaluation settings to produce an executable evaluation workflow, and finally generates task-oriented evaluation results and reports that support downstream decision making. A human-in-the-loop mechanism is integrated throughout the pipeline, allowing users to inspect, refine, and validate intermediate decisions when necessary.

At a high level, One-Eval consists of the following components. 
(1) \textbf{NL2Bench} translates natural language evaluation requirements into structured evaluation intents and recommends suitable benchmarks that align with user goals.
(2) \textbf{Benchmark Resolution and Configuration} completes dataset acquisition, configuration construction, and validation to ensure the evaluation workflow is executable and consistent.
(3) \textbf{Metric Recommendation and Reporting} selects evaluation metrics based on task requirements and produces structured, task-oriented evaluation reports rather than isolated scalar scores.

By explicitly modeling evaluation intent, workflow construction, and result interpretation as interconnected stages, One-Eval bridges the gap between user goals and executable evaluation pipelines. This design enables flexible customization, reduces manual configuration effort, and provides evaluation outputs that are directly actionable in practical deployment scenarios.

\subsection{NL2Bench}

NL2Bench is the entry point of One-Eval. Given a natural language evaluation request, it produces an executable \emph{benchmark plan}: a curated set of benchmarks together with the minimal metadata needed for downstream execution (e.g., canonical identifiers, evaluation splits, and schema hints). The plan can be iteratively refined through lightweight user interaction to ensure that the selected benchmarks truly match the user's intent.

\textbf{Intent Structuring.}
NL2Bench first translates the user request into a structured intent representation that captures (i) the target evaluation domain and capability focus (e.g., mathematical reasoning, general knowledge, text QA), (ii) any benchmarks explicitly specified by the user, (iii) execution constraints such as language or formatting requirements, and (iv) additional preferences that are difficult to encode as fixed fields. This structured representation serves as the control signal for subsequent retrieval and selection.

\textbf{Candidate Retrieval.}
Based on the structured intent, NL2Bench retrieves benchmark candidates from two complementary sources. The first source is a \emph{local benchmark gallery} of 77 curated benchmarks. We construct this gallery by collecting publicly available evaluation datasets, removing all entries whose data files cannot be successfully loaded or parsed, and retaining only those benchmarks that execute end-to-end without error. Each surviving benchmark is stored together with its canonical metadata (aliases, category tags, task-type annotations, HuggingFace configuration, and key mappings), forming a self-contained registry of ready-to-run evaluations. To match the user query against this gallery, we provide two interchangeable retrieval backends that share the same API: (i) an \emph{embedding-based} mode that encodes both the query and benchmark descriptions into dense vectors and ranks candidates by cosine similarity, and (ii) a lightweight \emph{TF-IDF} mode that tokenizes mixed Chinese--English text and combines cosine similarity with a keyword-overlap bonus, requiring no external service. A relevance threshold $\tau$ (set to 0.5 for embedding retrieval and 0.3 for TF-IDF) partitions the results into \emph{quality matches} and \emph{marginal matches}: when the number of quality matches is below the desired count $k$, the system falls back to a second source---live search over the HuggingFace Hub---to cover long-tail and newly released benchmarks. The threshold is calibrated so that the embedding mode, which produces semantically grounded similarity scores, applies a stricter cutoff to maintain precision, while the TF-IDF mode, whose scores are inherently noisier due to surface-level lexical matching, uses a more permissive cutoff to preserve recall. Candidates from both sources are merged with any user-specified benchmarks to form a unified pool for validation and selection.

\textbf{Resolution and Normalization.}
To ensure executability, NL2Bench normalizes each candidate into a canonical benchmark identifier and collects essential structural metadata. For external benchmarks, the agent reads dataset metadata (e.g., dataset cards and split/configuration information) and inspects feature fields when necessary, converting heterogeneous representations into a unified internal schema. Resolved benchmarks are presented in a benchmark gallery, which simultaneously provides user-facing explanations (why a benchmark is suggested) and supplies consistent configuration entry points for downstream execution.

\textbf{Selection Under Constraints.}
NL2Bench selects a compact subset of benchmarks that best match the user intent while respecting practical constraints such as evaluation cost, redundancy, and executability. In practice, this is implemented by combining intent-alignment scoring with rule-based validation, successful resolution checks, and budget-aware pruning. This design avoids over-selecting similar benchmarks and reduces the risk of producing plans that cannot be executed due to missing splits, incompatible schemas, or unavailable resources.

\textbf{Human-in-the-Loop.}
Because benchmark selection is inherently open-ended and misalignment can invalidate evaluation results, NL2Bench integrates human-in-the-loop refinement via interrupt points. The system shows the current benchmark plan with concise justifications (e.g., domain match, capability coverage, dataset characteristics) and allows the user to approve, edit the plan, refine the request, or inject a custom local benchmark. If the user modifies the intent, NL2Bench re-runs retrieval and selection until the user confirms a satisfactory plan.

The final output of NL2Bench is a user-approved benchmark plan with normalized identifiers, structural metadata, and configuration entry points, which is directly consumed by the next stage for executable resolution and configuration.

\subsection{Benchmark Resolution and Configuration}

Benchmark Resolution and Configuration, orchestrated by \texttt{BenchResolveAgent}, turns the nominal benchmark plan from NL2Bench (user-specified and recommended) into executable and reproducible configurations. To handle real-world heterogeneity in hosting sources, schemas, task definitions, and split conventions, the agent automatically resolves benchmark identifiers, acquires datasets when needed, and constructs validated configuration objects, enabling downstream evaluation to run without manual setup.

\paragraph{Hierarchical Benchmark Resolution.}
To balance stability for widely used benchmarks with extensibility to long-tail benchmarks, One-Eval adopts a hierarchical resolution strategy with a \emph{local-first, dynamic fallback} design. The system maintains a local registry of high-frequency benchmarks, each associated with expert-validated configurations. When a benchmark matches the registry, \texttt{BenchResolveAgent} loads the predefined configuration directly (including verified evaluation splits, column mappings, and task annotations), ensuring stable and reproducible execution across environments.

For benchmarks not found in the local registry, One-Eval falls back to HuggingFace for dynamic resolution: it first tries direct loading via the given name, and otherwise searches for candidates and selects the best match using lightweight heuristics (e.g., suffix cues and semantic similarity). Once resolved, the dataset and metadata are downloaded and integrated automatically, enabling seamless use of previously unseen community benchmarks without manual access or compatibility handling.

\paragraph{Unified Configuration and Heterogeneous Data Adaptation.}
To decouple evaluation logic from data representations, One-Eval normalizes each resolved benchmark into a unified configuration object (\texttt{BenchInfo}) stored in the system state. \texttt{BenchInfo} records the dataset source (HuggingFace ID or local path), the evaluation subset/split, a column mapping to One-Eval's standardized input--output interface, and task metadata for downstream metric recommendation. BenchResolve validates these fields during resolution and persists them as traceable artifacts (e.g., resolved IDs and cache paths), making protocol choices inspectable and reproducible across runs. This abstraction separates evaluation execution from data heterogeneity and enables seamless integration of curated internal benchmarks and community datasets, supporting scalable evaluation workflows in industrial settings.

\subsection{Metric Recommendation and Reporting}
Following the execution phase, this module serves as the analytical core, transforming the raw model outputs into actionable decision signals. Addressing the static evaluation framework and limited guidance (as highlighted in Sec.~\ref{sec:intro}), One-Eval adopts an agentic pipeline that couples semantic reasoning with rule-based priors to orchestrate metric selection, execution, and root-cause reporting.

\textbf{Dual-Track Metric Recommendation.}
To reconcile the flexibility required for unseen agentic tasks with the robustness needed for standard benchmarks, the \texttt{MetricRecommendAgent} implements a prioritized dual-track strategy that eliminates the need for manual configuration: (1) \textit{User Override (Static Control)}: explicit metric configurations provided in benchmark metadata take strict precedence, enabling bespoke evaluation protocols when required. (2) \textit{Knowledge-Augmented Reasoning (Dynamic Adaptation)}: for unconfigured or open-ended tasks, the agent performs semantic reasoning over rich dataset context (e.g., prompt templates, few-shot samples, task descriptors), grounded by \textit{dynamic prompt construction} that scans the registered metric library at runtime to generate semantic descriptions and decision rules; these are injected into the LLM context to guide metric selection.
% (e.g., \texttt{pass@k} for code generation vs. \texttt{exact\_match} for QA).
(3) \textit{Registry Fallback}: if the LLM fails to produce a valid plan, the system reverts to rule-based suggestions from the \texttt{MetricDispatcher} or a minimal default set to guarantee pipeline continuity.

\textbf{Decentralized Metric Registration.}
One-Eval provides an extensible metric ecosystem via a decentralized registration interface. New metrics are integrated by decorating computation functions with semantic metadata, after which the system automatically registers them into the global metric registry. This indexed library serves as the knowledge base for the agent’s recommendations.

\textbf{Execution Engine.}
Once metrics are selected, the \texttt{ScoreCalcAgent} invokes the \texttt{MetricRunner} as a unified execution layer. It normalizes heterogeneous inputs, aligns predictions with references, supports parallel execution for large-scale datasets, and packages results with scores, priorities, and details when available.

\textbf{Hierarchical Diagnostic Reporting.}
To overcome the limitation of isolated scalar metrics, One-Eval generates multi-granular diagnostic reports via \texttt{ReportGenAgent}: (1) \textit{Macro View (Capability Profiling)}: aggregates results into radar and sunburst summaries for holistic capability profiling. (2) \textit{Diagnostic View (Root Cause Analysis)}: attributes failure modes (e.g., instruction-following errors vs. hallucinations), performs blind-spot analysis over failed samples, and summarizes length distributions for correct vs. incorrect outputs. (3) \textit{Micro View (Case Study)}: provides case-level inspection tables that link aggregate metrics to specific failure instances.

\textbf{Specialized Metrics.}
To support the hierarchical reporting described above, One-Eval incorporates a comprehensive library of custom metrics designed to uncover specific failure modes. Table~\ref{tab:metric_taxonomy} highlights a representative subset of these featured metrics, selected to demonstrate how the system moves beyond standard accuracy to capture domain-specific nuances (e.g., symbolic equivalence in math) and behavioral patterns (e.g., format compliance). These metrics serve as the building blocks for the diagnostic views in the final report.

\section{Experiments}

We evaluate One-Eval from an industrial usability and reliability perspective. Rather than targeting leaderboard improvements on a fixed benchmark suite, our experiments focus on whether One-Eval can (i) produce actionable end-to-end evaluation outputs from natural-language requests with minimal user effort, (ii) reliably generate executable evaluation plans and run them through to results without human edits, and (iii) provide practical capabilities beyond existing evaluation frameworks. To this end, we conduct three complementary studies: (1) qualitative case studies that illustrate the full workflow and decision-oriented reporting, (2) a controlled end-to-end success-rate evaluation on a diverse set of evaluation requests to quantify executability and automation reliability, and (3) a feature-level comparison table against representative evaluation frameworks.

\subsection{Case Study}
\label{sec:case_study}

We present a real run to illustrate how One-Eval turns a single natural-language request into an executable and auditable evaluation workflow, and how it preserves an evidence trail for diagnosing failures.

\textbf{Request $\rightarrow$ plan.}
The user request is:
\vspace{-0.5em}
\begin{quote}
\small \emph{``I want to focus on broad general-knowledge coverage, and check whether the model can handle some light reasoning.''}
\end{quote}
\vspace{-0.7em}
One-Eval structures the request into intent slots (e.g., \texttt{domain=[text, reasoning]}) and uses NL2Bench to propose a benchmark plan spanning knowledge and reasoning, including \texttt{mmlu}, \texttt{truthful\_qa}, \texttt{commonsenseqa}, and reasoning-oriented sets such as \texttt{openai/gsm8k} and \texttt{HuggingFaceH4/MATH-500}. Non-canonical names are resolved to concrete repositories (e.g., \texttt{mmlu} $\rightarrow$ \texttt{cais/mmlu}) to ensure executability.

\textbf{Resolution, configuration, and schema normalization.}
BenchResolve automatically selects runnable configurations and splits with recorded rationales, and caches resolved datasets with reproducible paths. For example, GSM8K is configured as \texttt{\{config=main, split=test\}}, MATH-500 uses \texttt{\{config=default, split=test\}} (only available subset), and TruthfulQA falls back to \texttt{split=validation} (no \texttt{test} split). BenchResolve then normalizes heterogeneous schemas via key mappings (e.g., GSM8K maps \texttt{question} to input and \texttt{answer} to target; TruthfulQA maps \texttt{question} to input and \texttt{correct\_answers} to targets), enabling a unified evaluation interface across heterogeneous benchmarks.

\subsection{End-to-End Success Rate}

A central question for any automated evaluation system is whether it can turn a free-form requirement into a ready-to-run evaluation plan \emph{without human edits}. We collect 100 natural-language evaluation requests that span six broad capability domains (reasoning, mathematics, code, safety, retrieval, and factual QA) and feed each request into the One-Eval pipeline, running it from intent parsing through benchmark retrieval, configuration inference, and dataset preparation—stopping right before actual model evaluation. No manual correction is applied at any stage. We measure three cumulative success metrics along the pipeline and report the average decision time. Table~\ref{tab:e2e-success} summarizes the results.

\begin{table}[t]
\centering
\small
\caption{End-to-end success rates and efficiency on 100 evaluation requests. Each metric is cumulative: later checkpoints require all earlier ones to have succeeded.}
\label{tab:e2e-success}
\begin{tabular}{lcc}
\toprule
\textbf{Metric} & \textbf{Count} & \textbf{Rate} \\
\midrule
Plan Executable Rate  & 99/100  & 99\% \\
Auto-Complete Rate    & 85/100  & 85\% \\
Full Plan Rate        & 84/100  & 84\% \\
\midrule
\textbf{Avg. Tokens} & \multicolumn{2}{c}{10,652\ \;} \\
% \textbf{Med. Decision Time} & \multicolumn{2}{c}{686.96\,s\;($\approx$11.4\,min)} \\
\bottomrule
\end{tabular}
\vspace{-1em}
\end{table}

\noindent\textbf{Plan Executable Rate} measures whether the system can parse the user intent and retrieve at least one benchmark candidate without interruption; 99 out of 100 requests pass this checkpoint, with the single failure caused by a highly ambiguous query that the intent-structuring agent cannot decompose. \textbf{Auto-Complete Rate} further requires that the inferred subset, split, and key mappings are correct—errors at this stage would cause the downstream runner to crash or produce meaningless results; 85\% of requests clear this bar, indicating that the automated schema inference is reliable for the majority of publicly available benchmarks. \textbf{Full Plan Rate} additionally demands successful task-type inference and metric recommendation, reaching 84\%. The 8-step pipeline completes in a median wall-clock time of approximately 11.4 minutes per request (mean 13 minutes), demonstrating that One-Eval can deliver an executable evaluation plan from a single sentence of natural language within a practical time budget and without any human intervention.

\begin{table}[h]
\centering
\small
\setlength{\tabcolsep}{3pt}
\renewcommand{\arraystretch}{1.05}
\caption{Feature-level comparison.}
\vspace{-0.7em}
\label{tab:feature_comp}
\resizebox{\columnwidth}{!}{%
\begin{tabular}{lcccc}
\toprule
\textbf{Framework} & \textbf{Custom} & \textbf{Automate} & \textbf{Rec. Bench} & \textbf{Rec. Metric} \\
\midrule
One-Eval (ours)   & \cmark & \cmark & \cmark & \cmark \\
lm-eval-harness   & \cmark & \xmark & \xmark & \xmark \\
OpenCompass       & \cmark & \xmark & \xmark & \xmark \\
HELM              & \xmark & \xmark & \xmark & \xmark \\
\bottomrule
\end{tabular}%
}
\vspace{-1em}
\end{table}

\subsection{Feature-level Comparison}
\label{sec:feature_comp}

Table~\ref{tab:feature_comp} summarizes a feature-level comparison with representative evaluation frameworks. We focus on workflow-critical capabilities that affect practical benchmark evaluation, including customization, end-to-end automation, and intent-conditioned recommendations.

\section{Conclusion}
One-Eval enables natural-language evaluation requests to be executed as traceable end-to-end workflows via NL2Bench, BenchResolve, and task-aware Metrics \& Reporting. Experiments demonstrate reliable execution with minimal manual effort and actionable evidence for auditing and debugging. Future work will broaden coverage to more tasks and modalities and further strengthen support for long-tail benchmarks.

% \section*{Acknowledgments}

% This document has been adapted
% by Steven Bethard, Ryan Cotterell and Rui Yan
% from the instructions for earlier ACL and NAACL proceedings, including those for
% ACL 2019 by Douwe Kiela and Ivan Vuli\'{c},
% NAACL 2019 by Stephanie Lukin and Alla Roskovskaya,
% ACL 2018 by Shay Cohen, Kevin Gimpel, and Wei Lu,
% NAACL 2018 by Margaret Mitchell and Stephanie Lukin,
% Bib\TeX{} suggestions for (NA)ACL 2017/2018 from Jason Eisner,
% ACL 2017 by Dan Gildea and Min-Yen Kan,
% NAACL 2017 by Margaret Mitchell,
% ACL 2012 by Maggie Li and Michael White,
% ACL 2010 by Jing-Shin Chang and Philipp Koehn,
% ACL 2008 by Johanna D. Moore, Simone Teufel, James Allan, and Sadaoki Furui,
% ACL 2005 by Hwee Tou Ng and Kemal Oflazer,
% ACL 2002 by Eugene Charniak and Dekang Lin,
% and earlier ACL and EACL formats written by several people, including
% John Chen, Henry S. Thompson and Donald Walker.
% Additional elements were taken from the formatting instructions of the \emph{International Joint Conference on Artificial Intelligence} and the \emph{Conference on Computer Vision and Pattern Recognition}.

\bibliography{custom}
\clearpage
\newpage
\appendix
\label{sec:appendix}

\section{Alignment with Representative Evaluation Frameworks}
\label{sec:appendix_alignment}

This section provides additional details on how we align the criteria in Table~\ref{tab:feature_comp} with how representative evaluation frameworks are typically used in practice. Our goal is to make the comparison reproducible and interpretable by clarifying what each feature flag captures at the \emph{workflow} level.

\paragraph{Scope of the comparison.}
The feature-level comparison is intentionally scoped to capabilities that directly affect evaluation workflows in industrial settings: (i) integrating organization-specific assets (custom benchmarks/metrics), (ii) reducing front-loaded setup work (automation), and (iii) supporting intent-conditioned selection (recommendations) rather than static or manual choices. The table is not intended to rank overall quality, speed, or benchmark coverage, but to summarize whether a framework exposes these capabilities as first-class components.

\paragraph{Criterion 1: Custom benchmarks and metrics.}
We consider \textbf{Custom} supported when a framework provides a documented and maintainable mechanism to:
(i) register a new benchmark/dataset (or adapter) and
(ii) attach custom evaluation logic or metrics,
without modifying core runner internals.
In practice, this includes extension points such as benchmark registries, task adapters, dataset loaders, and pluggable metric modules. We also treat ``custom'' as supported when users can incorporate local datasets or private benchmarks through a clear interface (e.g., path-based loading with schema adapters) and keep them as reusable assets in subsequent runs.

\paragraph{Criterion 2: End-to-end automation.}
We consider \textbf{Automate} supported when the system can start from a high-level evaluation requirement and produce a runnable evaluation plan with minimal manual specification of:
benchmark identifiers, split/config selection, schema mappings, and execution settings.
This criterion reflects whether the framework reduces the manual effort of ``finding the right benchmark and making it runnable'' rather than only providing an execution harness once the benchmark list and configuration are already known. In One-Eval, automation is realized as an agentic workflow that (a) converts intent to a benchmark plan, (b) resolves identifiers, (c) downloads assets, and (d) normalizes schemas into executable configurations, while recording intermediate artifacts for traceability.

\paragraph{Criterion 3: Benchmark recommendation.}
We consider \textbf{Rec. Bench} supported when the system provides an explicit recommendation component that proposes benchmarks conditioned on user intent and constraints, instead of relying on static suites or manual selection. Concretely, this includes (i) intent structuring (domain/capability/constraints), (ii) candidate retrieval over curated registries and/or external spaces, and (iii) producing a benchmark plan with justifications that can be inspected or refined. One-Eval implements this capability in NL2Bench, which retrieves candidates from both curated registries and open repositories and outputs an \emph{EvalPlan} that serves as the entry point for downstream execution.

\paragraph{Criterion 4: Metric recommendation.}
We consider \textbf{Rec. Metric} supported when the system proposes a metric suite (and corresponding reporting template) conditioned on the task type and evaluation objective, rather than always reporting a fixed scalar (e.g., accuracy) or requiring users to manually choose metrics for each benchmark. This criterion covers not only selecting a metric function, but also packaging metrics into a task-oriented report structure (e.g., breakdowns, slices, and diagnostic signals). In One-Eval, metric recommendation is coupled with report generation to support decision-oriented evaluation outputs.

\paragraph{Interpretation of blanks/\xmark.}
Blank/\xmark\ indicates that the capability is not exposed as an intent-conditioned, first-class module in the framework and is typically achieved through manual selection/configuration or external scripting around the runner. This does not preclude users from implementing similar behavior with sufficient engineering effort; rather, the table summarizes what is provided natively as part of the framework's workflow.

\paragraph{Reproducibility and reporting.}
To keep the comparison reproducible, we apply the above criteria consistently and focus on user-facing workflow primitives (configuration interfaces, registries, and built-in planning/recommendation modules). For One-Eval, the corresponding artifacts (plans, resolved benchmark identifiers, configuration choices, and report outputs) are persisted as part of the traceable evaluation state, allowing readers to map each checked feature to concrete system outputs.

\section{Selected Custom Metrics}
Table~\ref{tab:metric_taxonomy} summarizes a curated set of One-Eval's custom metrics designed for diagnostic insight. These metrics go beyond generic aggregate scores by targeting specific failure modes (e.g., symbolic equivalence, schema/format compliance, and judge-based error attribution), enabling more actionable analysis and decision-oriented reporting across tasks.
\begin{table*}[h]
\centering
\small
\caption{\textbf{Selected Custom Metrics for Diagnostic Insight.} This table presents a non-exhaustive list of featured metrics in One-Eval, categorized by their diagnostic focus. Unlike generic scalar metrics, these are engineered to pinpoint specific error types and behavioral anomalies.}
\label{tab:metric_taxonomy}
\begin{tabular}{c|c|m{9cm}} 
\toprule
\textbf{Category} & \textbf{Featured Metric} & \textbf{Diagnostic Utility \& Mechanism} \\
\midrule
\multirow{2}{*}[-1ex]{\textbf{Math Reasoning}} & \texttt{math\_verify} & \textbf{Hybrid Equivalence}: Combines strict text matching with mathematical equivalence checks to reduce false negatives in varying output formats. \\
 & \texttt{symbolic\_match} & \textbf{Algebraic Validation}: Uses symbolic simplification libraries to verify correctness regardless of variable ordering or simplification state. \\
\midrule
\multirow{2}{*}[-1ex]{\textbf{Code Generation}} & \texttt{soft\_code\_execution} & \textbf{Static Analysis}: Performs syntax parsing and complexity checks (e.g., cyclomatic complexity) without requiring a sandboxed runtime. \\
 & \texttt{code\_similarity} & \textbf{Reference Proxy}: Computes BLEU-based similarity against ground truth when executable test cases are unavailable. \\
\midrule
\multirow{3}{*}[-2.5ex]{\textbf{Behavioral}} & \texttt{format\_compliance} & \textbf{Instruction Adherence}: Quantifies the rate of successful structural parsing (e.g., JSON/Markdown), critical for downstream system integration. \\
 & \texttt{extraction\_rate} & \textbf{Answer Stability}: Measures the model's ability to isolate the final answer from its own reasoning chain (Chain-of-Thought). \\
 & \texttt{reasoning\_efficiency} & \textbf{Verbosity Check}: Penalizes excessive token usage that does not contribute to information gain, identifying "chatty" failure modes. \\
\midrule
\multirow{2}{*}[-1ex]{\textbf{LLM-based}} & \texttt{case\_study\_analyst} & \textbf{Error Attribution}: Auto-samples failed cases and classifies error types (e.g., Hallucination vs. Logic Error) using a judge model. \\
 & \texttt{gini\_index} & \textbf{Capability Balance}: Measures the evenness of performance across different task categories to detect domain overfitting. \\
\bottomrule
\end{tabular}
\end{table*}

\section{Additional Discussion}
\label{sec:appendix_discussion}

This section discusses practical considerations that motivate One-Eval's design, focusing on deployment-oriented evaluation and how decision-makers consume evaluation outcomes.

\paragraph{Decision-oriented evaluation outputs.}
In industrial settings, evaluation is commonly used to support decisions such as model selection, release gating, regression monitoring, and targeted iteration. As a result, practitioners often need more than an aggregate score: they need structured signals that help identify \emph{where} a model succeeds or fails and \emph{why} performance changes across versions. One-Eval operationalizes this by producing task-oriented reports that naturally accommodate multiple diagnostic views, such as breakdowns by benchmark/subset, failure-mode summaries, and objective-aligned indicators that can be reused in iterative evaluation cycles.

\paragraph{From scores to actionable signals.}
A practical report is most useful when it bridges model outputs to actions. One-Eval structures evaluation artifacts so that stakeholders can quickly connect (i) aggregate results to (ii) slices or categories of interest and (iii) representative examples. This structure supports common operational needs: identifying regressions, prioritizing data collection or prompt/template adjustments, and communicating evaluation findings across roles (engineering, product, and quality). By packaging these signals consistently, One-Eval supports evaluation as a repeatable process rather than an ad-hoc analysis step.

\paragraph{Traceability and auditability for iterative evaluation.}
Industrial evaluation often involves repeated runs over time, comparing model variants, configuration changes, and shifting requirements. One-Eval treats intermediate artifacts as first-class state (plans, configurations, resolved benchmark identifiers, schema mappings, caches, and per-sample traces), enabling results to be audited and reproduced. This is particularly important when evaluation outcomes must be explained to stakeholders or used as part of a release decision, where reproducible evidence reduces ambiguity and accelerates review.

\paragraph{Human-in-the-loop as controlled refinement.}
Evaluation involves intent-sensitive choices (e.g., what constitutes ``coverage'' for a domain) and configuration-sensitive choices (e.g., how to handle multiple subsets/splits). One-Eval integrates human-in-the-loop interaction as controlled refinement at key checkpoints: users can confirm benchmark coverage, adjust constraints, and approve reporting preferences. In typical usage, these interactions are lightweight and focus on validating high-impact decisions, ensuring that automation remains aligned with the user's evaluation objective.

\paragraph{Supporting evolving and heterogeneous requirements.}
Evaluation requirements vary across domains (reasoning, knowledge, safety, retrieval-augmented settings) and evolve rapidly as new benchmarks and practices emerge. One-Eval separates intent understanding, benchmark resolution/configuration, and evaluation/reporting into modular stages. This separation enables the system to incorporate new benchmark sources, add domain-specific report templates, and extend metric suites while keeping a stable end-to-end workflow abstraction for users.

\section{Consistency with Benchmark Cards and Standard Protocols}
\label{sec:appendix_consistency}

This section discusses how One-Eval promotes evaluation consistency by grounding configuration and execution in benchmark metadata (e.g., dataset cards) and by recording protocol-relevant decisions as reproducible artifacts.

\paragraph{Protocol grounding via benchmark metadata.}
Benchmarks in open repositories often provide guidance in dataset cards, including intended task definition, available configurations/subsets, supported splits, label semantics, and field descriptions. One-Eval retrieves and records such metadata during benchmark resolution and uses it as a primary signal for configuration planning. Concretely, metadata is used to (i) disambiguate benchmark identifiers, (ii) identify the canonical configuration when multiple subsets exist, and (iii) surface split availability and constraints that affect executability.
\vspace{-0.5em}
\paragraph{Split and subset selection policy.}
To align with standard protocols, One-Eval prioritizes evaluation on the canonical split when available (commonly \texttt{test}). When a \texttt{test} split is not provided, the system selects the best-supported alternative (commonly \texttt{validation}) and records the selection and rationale as part of the benchmark configuration state. For benchmarks with multiple subsets/configurations, One-Eval selects a representative subset using metadata cues (e.g., subset descriptions and common naming conventions) and records the resolved choice to ensure that subsequent runs follow the same protocol unless explicitly refined.
\vspace{-0.5em}
\paragraph{Versioned identifiers and reproducible artifacts.}
Consistency requires that reported results can be traced back to the exact evaluated data. One-Eval records resolved benchmark identifiers (e.g., repository IDs), configuration parameters (subset/config/split), and reproducible cache locations for downloaded artifacts. When supported by the underlying data source, One-Eval also records version-like information (e.g., dataset revision/commit or snapshot metadata) alongside cache paths, making it straightforward to rerun the same evaluation protocol across machines and over time.

\paragraph{Schema alignment with benchmark definitions.}
Benchmarks exhibit heterogeneity in feature schemas (e.g., \texttt{question}/\texttt{answer} vs.\ \texttt{instruction}/\texttt{output}, or single-target vs.\ multi-reference targets). One-Eval resolves these differences by inspecting dataset features and constructing explicit key mappings into a unified input--output interface. Importantly, these mappings are stored with the benchmark configuration and can be inspected or refined, so that schema normalization remains transparent and consistent with how the benchmark is defined.

\paragraph{Executability and validity checks.}
To ensure that protocol choices remain meaningful in execution, One-Eval performs lightweight validity checks during resolution and before evaluation runs, such as verifying the presence of required fields (inputs, targets/labels) and confirming that the selected split contains evaluable targets. These checks complement metadata grounding and help preserve consistent evaluation behavior across heterogeneous benchmarks.

\paragraph{Auditable evidence trails for standard consistency.}
For each run, One-Eval records protocol-relevant artifacts including the benchmark plan, final resolved benchmark set, per-benchmark configuration decisions, and per-sample traces. This artifact-level evidence trail supports auditability: readers and practitioners can verify what was evaluated under which protocol choices, and can reproduce results by reusing the recorded configurations and cache references.

\onecolumn            % <—— 暂时切回单栏
\section{Prompts}
\newcommand{\promptpagex}[1]{
  \begin{figure}[h]
    \centering
    \includegraphics[page=#1, width=0.9\textwidth, keepaspectratio]{latex/Imgs/prompt-eval.pdf}
  \end{figure}
}
% This section include these prompts: difficulty and category classification, data filtering, question synthesis, and correctness verification.
\promptpagex{1}

\newcommand{\promptpage}[1]{
  \begin{figure}[t]
    \centering
    \includegraphics[page=#1, width=0.9\textwidth, keepaspectratio]{latex/Imgs/prompt-eval.pdf}
  \end{figure}
}

\promptpage{2}
\promptpage{3}
\promptpage{4}
\promptpage{5}
\promptpage{6}
\promptpage{7}
\promptpage{8}
\twocolumn            % <—— 之后再回到双栏

\end{document}